\pdfoutput=1
\documentclass[runningheads]{llncs}
\usepackage{graphicx}

\usepackage{tikz}
\usepackage{comment}
\usepackage{amsmath,amssymb} 
\usepackage{color}

\usepackage{xspace}
\usepackage{amsmath}
\usepackage{amssymb}
\usepackage{booktabs}
\usepackage{placeins}
\usepackage{float}
\usepackage{subfloat}

\usepackage{algorithm}
\usepackage{algorithmicx}
\usepackage{algpseudocode}

\usepackage{todonotes}

\usepackage{booktabs}  
\usepackage{multirow}  

\usepackage{graphicx}
\usepackage{microtype}
\newcommand{\NA}{---}

\usepackage{subcaption}
\usepackage{bm} 
\usepackage{pifont}
\usepackage{color}

\newcommand{\sys}{\textsc{DisTrans}\xspace}
\newcommand{\trigger}{offset\xspace}

\usepackage{wrapfig}

\usepackage{hyperref}
\hypersetup{colorlinks}

\usepackage[capitalize]{cleveref}
\crefname{section}{Sec.}{Secs.}
\Crefname{section}{Section}{Sections}
\Crefname{table}{Table}{Tables}
\crefname{table}{Tab.}{Tabs.}

\newenvironment{icompact}{
  \begin{list}{$\bullet$}{
    \itemindent -.05em
    \parsep 0pt plus 1pt
    \partopsep 0pt plus 1pt
    \topsep 2pt plus 2pt minus 2pt
    \itemsep 0pt plus 1.3pt
    \parskip 0pt plus 2pt
    \leftmargin 0.13in}
      }
{\normalsize\end{list}}

\begin{document}
\pagestyle{headings}
\mainmatter
\def\ECCVSubNumber{6551}  

\title{Addressing Heterogeneity in Federated Learning via Distributional Transformation} 

\titlerunning{FL Distributional Transformation}
%
\author{
Haolin Yuan\inst{1}$^*$, 
Bo Hui\inst{1}$^*$,
Yuchen Yang\inst{1}$^*$,
Philippe Burlina\inst{1,2}, \\ 
Neil Zhenqiang Gong\inst{3}, 
 and
Yinzhi Cao\inst{1}} 


%
\authorrunning{H. Yuan et al.}
%
\institute{Department of Computer Science, Johns Hopkins University \\
\email{\{hyuan4, bo.hui, yc.yang, yinzhi.cao\}@jhu.edu}\\ \and
Johns Hopkins University Applied Physics Laboratory (JHU/APL)
\email{Philippe.Burlina@jhuapl.edu}\\ \and
Duke University\\
\email{neil.gong@duke.edu}}
\maketitle
\begingroup\renewcommand\thefootnote{$^*$}
\footnotetext{The first three authors have equal contributions to the paper. }
\endgroup

\begin{abstract}



Federated learning (FL) allows multiple clients to collaboratively train a deep learning model. One major challenge of FL is when data distribution is heterogeneous, i.e., differs from one client to another.  
%
 Existing personalized FL algorithms are only applicable to narrow cases, e.g., one or two data classes per client, and therefore they do not satisfactorily address FL under varying levels of data heterogeneity.  In this paper, we propose a novel 
 framework, called \sys, to improve FL performance (i.e., model accuracy) via train and test-time distributional transformations along with a double-input-channel model structure. 
  \sys works by optimizing distributional offsets and models for each FL client to shift their data distribution, and aggregates these offsets at the FL server to further improve performance in case of distributional heterogeneity. 
 %
  Our evaluation on multiple benchmark datasets shows that \sys outperforms state-of-the-art FL methods and data augmentation methods under various settings and different degrees of client distributional heterogeneity.

\end{abstract}

\section{Introduction}


Federated learning~\cite{FedAvg,FL_intro_1,FL_intro_2,FL_intro_3} (FL) is an emerging  distributed machine learning (ML) framework that enables clients to learn models together with the help of a central server. 
 In FL, each client learns a local model that is sent to the FL server for aggregation, and subsequently the FL server returns the aggregated model to the client.  The process is repeated until convergence. One emerging and unsolved FL challenge is that the data distribution at each client can be heterogeneous. For example, for FL based skin diagnostics, the skin disease distribution for each hospital / client can vary significantly. In another use case of smartphone face verification, data distributions collected at each mobile device can vary from one client to another. Such distributional heterogeneity often 
  leads to suboptimal accuracy of the final FL model.

  \begin{figure}[!t]
  \centering
  \includegraphics[width=1\linewidth]{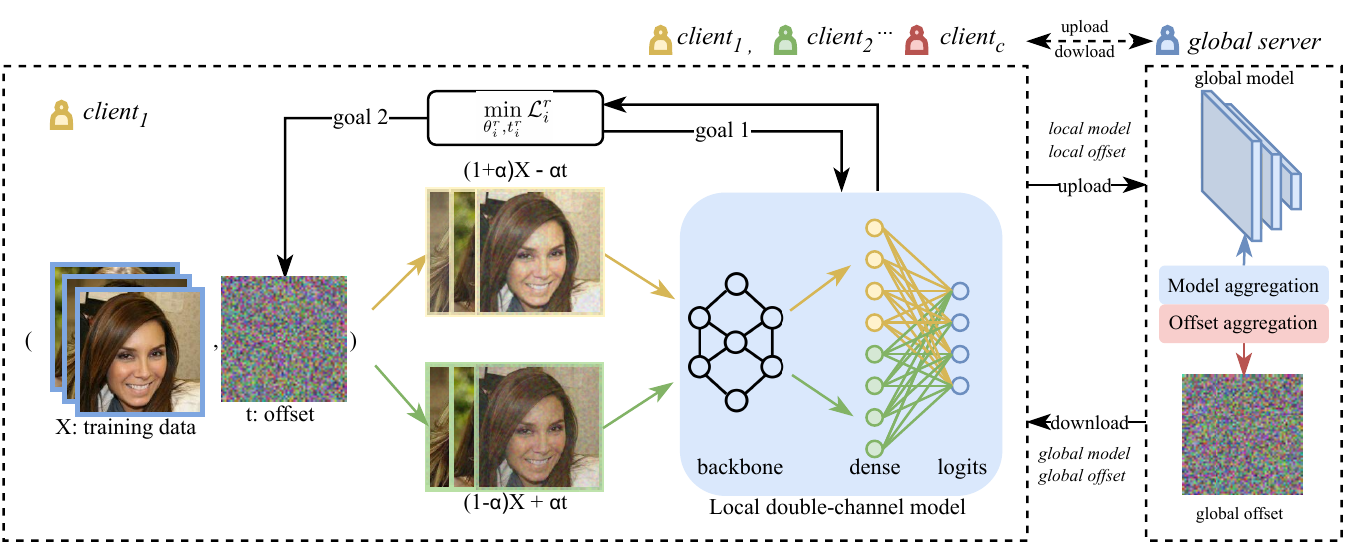} 
  \caption{The pipelines of \sys. Each client  jointly optimizes the offset and model in local training phase, then uploads both to the central server for aggregation. The aggregated model and offset are sent back to clients for next-round. } 
  \label{fig:arch}
  \end{figure}

There are two types of approaches to learn FL models under data heterogeneity: (i) improving FL's training process  and (ii) improving clients' local data. Unfortunately, neither improves FL under varied levels of data heterogeneity. 
 On one hand, existing FL methods~\cite{FedAvg,non_personalized_FL_1,non_personalized_FL_2}, especially personalized FLs~\cite{FL_different_model_1,FL_different_model_2}, learn a model (or even multiple models) using customized loss functions or model architectures based on heterogeneity level. However, existing personalized FL algorithms are designed for highly heterogeneous distribution. 
 %
 %
 %
  FedAwS~\cite{FedAws} can only train FL models when local client's data has one positive label. 
  The performance of pfedMe~\cite{pFedMe} and pfedHN~\cite{hypernet} degrades to even 5\% to 18\% lower accuracy than FedAvg~\cite{FedAvg}, when the data distribution is between heterogeneity and homogeneity. 
  

On the other hand, traditional centralized machine learning also rely on data transformations, i.e., data augmentation,~\cite{augment_1,augment_2,augment_3,augment_4,augment_5,augment_6,augment_7,augment_8} to improve model's performance.  Such transformations could be used for a pre-processing of all the training data or an addition to the existing training set. Until very recently, data transformations are also used during test time~\cite{tta-1,tta-2,tta-3,tta-4,tta-5} to improve learning models, e.g.,  adversarial robustness~\cite{tta-1}. However, it remains unclear whether and how data transformation can improve FL particularly under different client heterogeneity. The major challenge is how to tailor transformations for each client with different data distributions.

  In this paper, we propose the {\it first} 
  FL distributional transformation framework, called \sys, to address this heterogeneity challenge by altering local data distributions via a client-specific data shift applied both on train and test/inference data.  Our distributional transformation alters each client's data distribution so that such distribution becomes less heterogeneous and thus the local models can be better aggregated at the server. 
  %
  %
  %
   Specifically, \sys performs a so-called {\it joint optimization}, at each client, to train the local model and generate an offset that is added to the local data. That is, an \sys's client alternately performs two steps in each round: 1) optimizing the personalized \emph{offset} to transform the local data via distribution shifts and 2) optimizing a local model to fit its offsetted local data. After client-side optimization, the FL server aggregates both the personalized offsets and the local models from all the clients and sends the aggregated global model and offset back to each client. During testing, each client adds its personalized offset to each testing input before using the global model to predict its label. 
  


\sys is designed with a special network architecture, called a double-input-channel model, to accommodate client-side offsets. 
 This double-input-channel model has a backbone network shared by both channels, a dense layer accepting outputs from two channels in parallel, and a logits layer that merges channel-related outputs from the dense layer. This double architecture allows the offset to be added to an (training or testing) input in one channel but subtracted from the input in the other. Such addition and subtraction better preserves the information in the original training and testing data because the original data can be recovered from the data with offset in the two channels.

  %
  
  We perform extensive evaluation of \sys using five different image datasets and compare it against state-of-the-art (SOTA) methods. Our evaluation shows that \sys outperforms SOTA FL methods across various  distributional settings of the clients' local data by 1\%--10\% with respect to testing accuracy. Moreover, our evaluation shows that \sys achieves 1\%--7\% higher testing accuracy than other data transformation / augmentation approaches, i.e.,  mixup~\cite{mixup} and AdvProp~\cite{augment_3}. 
  The code for \sys is made available under (\url{https://github.com/hyhmia/DisTrans}).

\section{Related Work}

 
 Existing federated learning (FL) studies focus on improving accuracy~\cite{FedAvg,FedAws,pFedMe,hypernet}, convergence~\cite{FL_convergence_3,FL_convergence_2,FL_convergence_1,FL_convergence_4,FL_convergence_5,FL_convergence_6}, communication cost~\cite{FL_cost_STOA_1,FL_cost_STOA_2,FL_cost_STOA_3,FL_cost_STOA_4,FL_cost_5,FL_cost_4,FL_cost_1,FL_cost_2}, security and privacy~\cite{federated-active,fang2020local,cao2021provably,cao2020fltrust}, or others~\cite{FL_derived_5,FL_derived_1,FL_derived_4,FL_derived_2}. Our work focuses on FL accuracy.

\noindent{\bf Personalized Federated Learning.} Prior studies~\cite{pFedMe,FedAws,hypernet} have attempted to address personalization, i.e., to make a model better fit a client's local training data.  For instance, FedAwS~\cite{FedAws} investigates  FL problems where each local model only has access to the positive data associated with only a single class and imposes a geometric regularizer at the server after each round to encourage classes to spread out in the embedding space.
 pFedMe~\cite{pFedMe} formulates a new bi-level optimization problem and uses Moreau envelopes to regularize each client  loss function and to decouple personalized model optimization from the global model learning. pFedHN~\cite{hypernet} utilizes a hypernetwork model as the global model to generate weights for each local model. MOON~\cite{moon} uses contrastive learning to maximize the agreement between local and global model.

\noindent{\bf Data Transformation.} Data transformation applies label-preserving transformations to images and is a standard technique to improve model accuracy in centralized learning. Most
of the recent data transformation methods~\cite{augment_1,augment_2,augment_3,augment_4,augment_5,augment_6,augment_7,augment_8} focus on transforming datasets during
the training phase. For instance, mixup~\cite{mixup} transforms the training data by mixing up the features and their corresponding labels; and AdvProp~\cite{augment_3} transforms the training data by adding adversarial examples. Additionally, transforming data at testing time~\cite{tta-1,tta-2,tta-3,tta-4,tta-5} has received increased attention. The basic test-time transformations use multiple data augmentations~\cite{tta-4,tta-5} at test time to classify one image and get the averaged results. P{\'e}rez et.al~\cite{tta-1} aims to enhance adversarial robustness via test-time transformation. As a comparison, \sys is the first to utilize test-time transformation to  improve federated learning accuracy under data heterogeneity.

\section{Motivation}


\begin{figure*}[!t]
    \subcaptionbox{$y=cos(wx)$ on  local clients.\label{fig:moti1}}{\includegraphics[trim=0 10 0 0, width=0.46\linewidth]{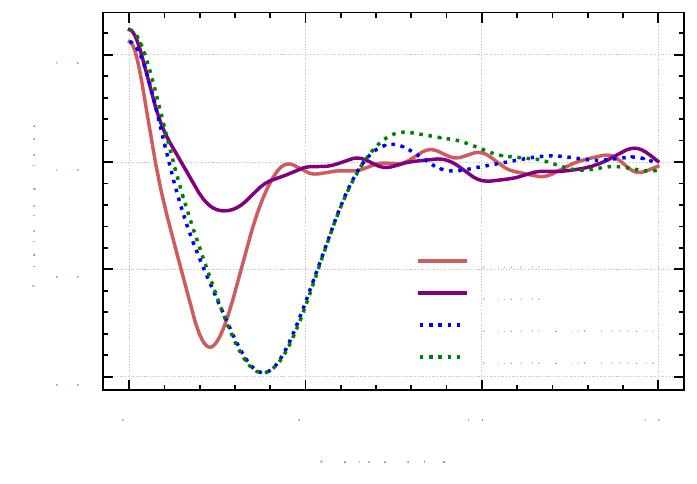}} \hfill
    \subcaptionbox{$y=wx$ on FL clients.\label{fig:moti3} }
    {\includegraphics[trim=0 10 0 0, width=0.46\linewidth]{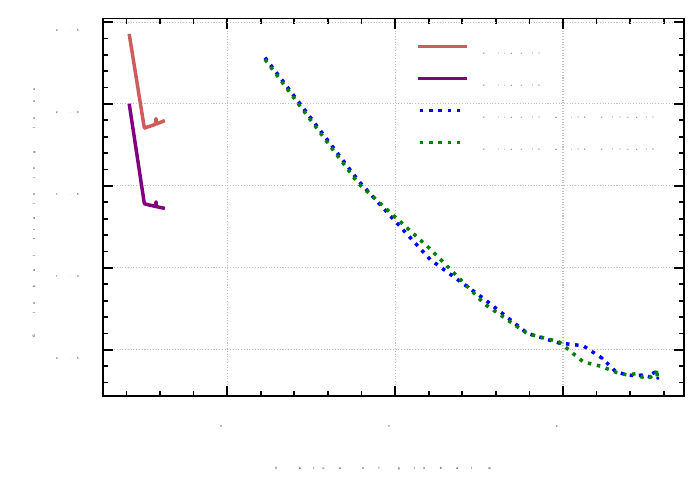}} 
    \caption{Training loss with respect to optimal weight $w$ on two clients' local training data with and w/o \trigger. We observe that offsets can make the training loss against weight more consistent on local clients and help FL model converge.} 
    \label{fig:moti}
\vspace{-15pt}
\end{figure*}

\sys's intuition is to transform each client's training and testing data with offsets to improve FL under heterogeneous data. That is, \sys transforms the client-side data distribution so that the learned local models are less heterogeneous and can be better aggregated. 
 To better illustrate this intuition, we describe two simple learning problems as motivating examples. Specifically, we show that well-optimized and selected offsets can (i) align two learning problems at different FL clients and (ii) help the aggregated model converge. 

\noindent{\bf Local Non-convex Learning Problems.} We consider a non-convex learning problem, i.e., $f(x)=cos(wx)$ where $w \in \mathbb{R}$, at two local clients with heterogeneous data.  The local data is generated via $x,y\in \mathbb{R}$ with $y=cos(w_{clientk}^{true} x)+\epsilon_{clientk}$, where $x$ is drawn i.i.d from Gaussian distribution and $\epsilon_{clientk}$ is Gaussian noise with mean value as 0. 
The offsets are
 $p x + q$ where $p$ is a fixed value at both clients and $q$ is chosen via brute force search. 
 Figure~\ref{fig:moti1} shows the squared training loss with and without offsets. 
  The difference between the training losses of two learning models are reduced, thus making two clients consistent. 

\noindent{\bf Linear Regression Problems with An Aggregation Server.} We train two local linear models, i.e.,  $f(x)=wx$ with the model parameter $w \in \mathbb{R}^2$,  aggregate the parameters at a server following FL, and then repeat the two steps following FL until convergence. The local training data is heterogeneous and generated as
  $y=w^{true}_{clientk}x+\epsilon_{clientk}$, where each of the two dimensions of $x$ is drawn i.i.d from normal distribution and $\epsilon_{clientk}$ is a Gaussian noise. 
 The offset is the same as the non-convex learning problem. We fix $p$ and optimize $q$ and $w$ via SGD at each client to minimize learning loss respectively. 
  Figure~\ref{fig:moti3} shows the squared training loss with respect to the optimal $w$ (sum value of two dimensions) with and without the offsets. Clearly, when offsets are not present, the aggregated model does not converge, resulting in a set of sub-optimal weights.  Instead, the aggregated model converges with a small training loss with the presence of offsets, confirming our intuition.

\section{Method}

In this section, we present our proposed method in detail. \sys aims to learn a single shared global model for the clients. Algorithm~\ref{algo-1} shows the pseudo-code of \sys. In each round, each client learns a local model and an offset, which are sent to the central server. The server aggregates the clients' local models and local offsets, and sends them back to the clients.  
Based on the intuition presented above, we propose a joint optimization method to learn a local model and offset for a client in each round.  
Figure~\ref{fig:arch} illustrates our joint optimization that each client performs.

\renewcommand{\algorithmicrequire}{\textbf{Input:}}
\renewcommand{\algorithmicensure}{\textbf{Output:}}
    \begin{algorithm}[!t] \scriptsize
        \caption{Pseudo-code of \sys}
        \label{algo-1}
        \begin{algorithmic}[1] 
            \Require  Number of clients $C$, local training dataset $D_i$ for client $i$, number of rounds $R$, batch size $B$, number of epochs $E$, and learning rates $\eta$ and $\eta_p$ for  model and offset $t$, respectively
            \Ensure Offset $t_i$ for client $i$ and global model $\theta$
                \State Server initializes global model $\theta^0$ and offset $t_i^0$ for each client $i$
                \For{$r = 0$ to $R-1$}
                    \State Server sends $\theta^r$ and $t^r_i$ to client $i$
                    \For{$i = 0$ to $C-1$}
                            \State $\theta^r_i \gets \theta^r$
                            // Initialize local model $\theta^r_i$ for client $i$
                            \For{$e = 0$ to $E-1$}
                            \For{each mini-batch $D_m$ from $D_i$}

                               \State $t^r_i \gets SGD(\nabla_{t^r_i}\mathcal{L}_i^r, t^r_i, \eta_t)$
                                // Update offset $t^r_i$
                                 
                                \State $x_t \gets ((1- \alpha)  x +  \alpha  t^r_i,\ (1+ \alpha)  x - \alpha  t^r_i )$
                                // Combine $t^r_i$ with each $x\in D_m$
                            
                                \State $\theta^r_i \gets SGD(\nabla_{\theta^r_i} \mathcal{L}_i^r,{\theta^r_i} , \eta)$
                                // Update local model 
                            \EndFor
                            \EndFor
                            \State Client $i$ sends $\theta^r_i$ and $t^r_i$ to server
                    \EndFor
                    \State  Server updates global model: $\theta^{r+1} \gets \frac{1}{C} \sum_{i\in[C]}\theta^r_i$
                    \State Server updates offset $t^{r+1}_i$ for each client $i$ via Offset Aggregation
                \EndFor
        \end{algorithmic}
    \end{algorithm}


\noindent{\bf Notations.} We assume $C$ clients and denote by $D_i$     the local training dataset for client $i$, where $i=1,2,\cdots,C$ and $|D_i|= n_i$. We consider $z = (x,y)$  a training sample, where $x \in \mathbb{R}^m$ denotes the training input and $y$ the label of the training input. We also denote by $D_{ti}$ the offsetted local training dataset for  client $i$, $x_t$ an offsetted training input, and $z_t=(x_t, y)$ a training sample offsetted with  \trigger.  We denote by $\theta$ the global model. 


\subsection{{Double-Input-Channel} Model Architecture}


\sys uses a double-input-channel neural network architecture (see Figure~\ref{fig:arch}) for a local/global model. Our architecture has a shared backbone network, a dense layer concatenating two channels' outputs, and a logits layer merging outputs from the dense layer.  Specifically, these two channels shift the local data distribution in two different ways using the same offset $t$. Formally, Eq.~\ref{blending} shows our two linear shifts:
\begin{equation} 
x_t = ((1- \alpha)  x +  \alpha  t,\ (1+ \alpha)  x - \alpha  t )),
\label{blending}
\end{equation}
where the first channel adds the offset $t$ to the input $x$ with a coefficient $\alpha$ (i.e., $(1- \alpha)  x +  \alpha  t$ is the input for the first channel) and the second subtracts $t$ from $x$ with the same $\alpha$ (i.e., $(1+ \alpha)  x - \alpha  t $ is the input to the second channel). Unless otherwise mentioned, our default setting for $\alpha$ is 0.3 in our experiments.

\subsection{Joint Optimization}

 In our joint optimization, each client aims to achieve the following two goals:
 \begin{icompact}
\item {\em Goal 1.} Optimizing {\it \trigger} to shift local data distribution to better fit with local model.
\item {\em Goal 2.} Optimizing local model to fit with offsetted local data distribution.
\end{icompact}

We formulate the two goals as an optimization problem. Specifically, client $i$ aims to solve the following optimization problem in  round $r$:
\begin{equation}  
\min_{\theta_i^r, t_i^r} \mathcal{L}_i^r = \frac{1}{n}\sum_{z_t\in D_{ti}} l(\theta_i^r, z_t),
\label{equ:loss}
\end{equation}
where $\theta_i^r$ is the local model of client $i$, $t_i^r$ is the \trigger of client $i$, and $\mathcal{L}_i^r$ is the loss function of client $i$ in round $r$. We choose cross entropy as loss term in our implementation. Solving $t_i^r$ in Eq.~\ref{equ:loss} while fixing $\theta_i^r$ achieves Goal 1; and solving $\theta_i^r$ in Eq.~\ref{equ:loss} while fixing $t_i^r$ achieves Goal 2. Therefore, we initialize $\theta_i^r$ as the global model $\theta^r$ and alternately optimize $t_i^r$ and $\theta_i^r$ for each mini-batch. Algorithm~\ref{algo-1} illustrates our pseudo-code.

\subsection{Model and Offset Aggregation}

The server aggregates both the local models and the offsets from the clients. The model aggregation follows the traditional FL, e.g., the server computes the mean of the clients' local models as the global model like FedAvg~\cite{FedAvg}. Our offset aggregation leverages the class distribution at each client.  Next, we first introduce a metric to measure \emph{distributional heterogeneity} and then our offset aggregation method  based on the metric. 


\begin{figure*}
    \subcaptionbox{{$DH=0\%$} }{\includegraphics[width=0.245\linewidth]{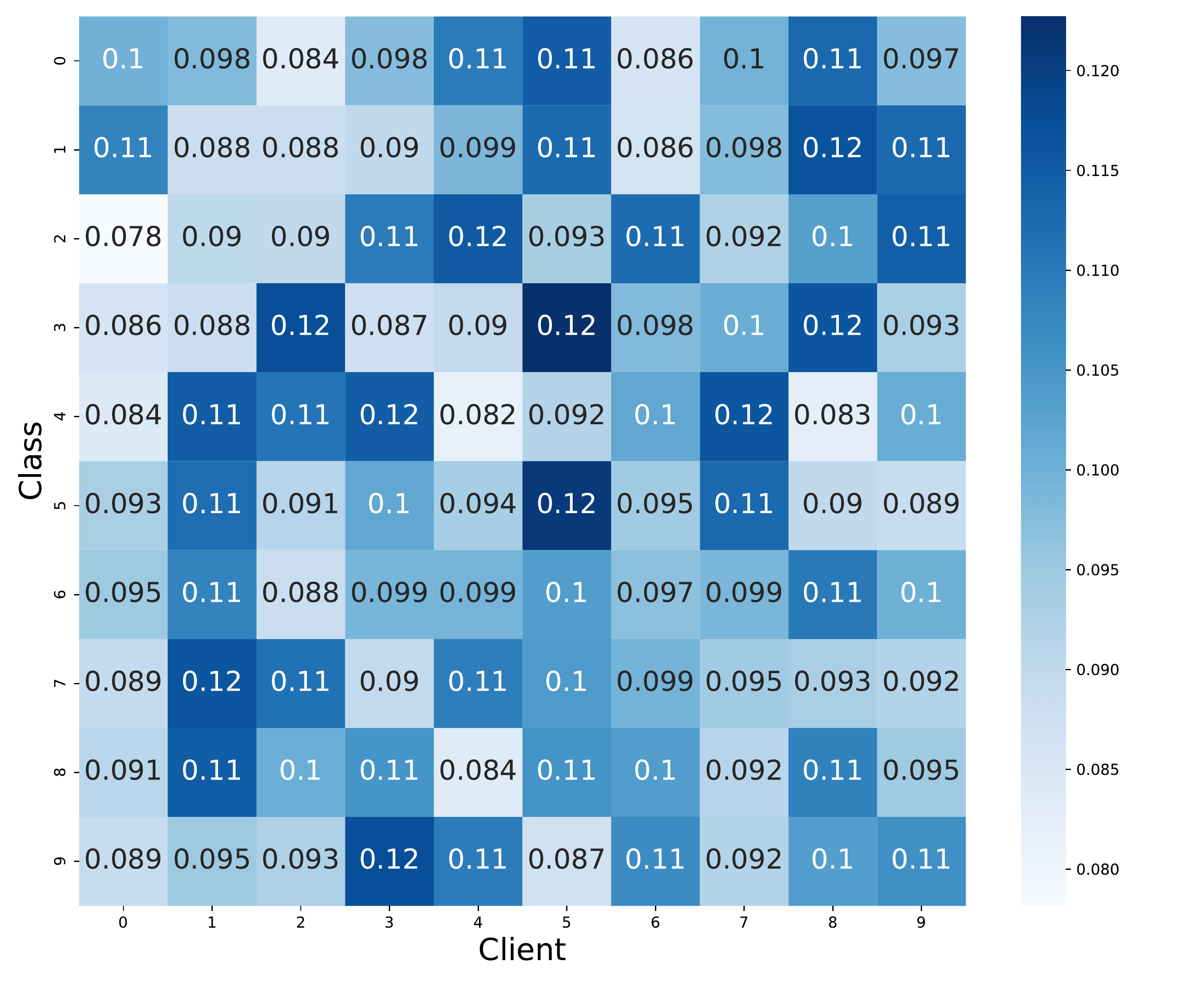}}
    \subcaptionbox{$DH=40\%$ }{\includegraphics[width=0.245\linewidth]{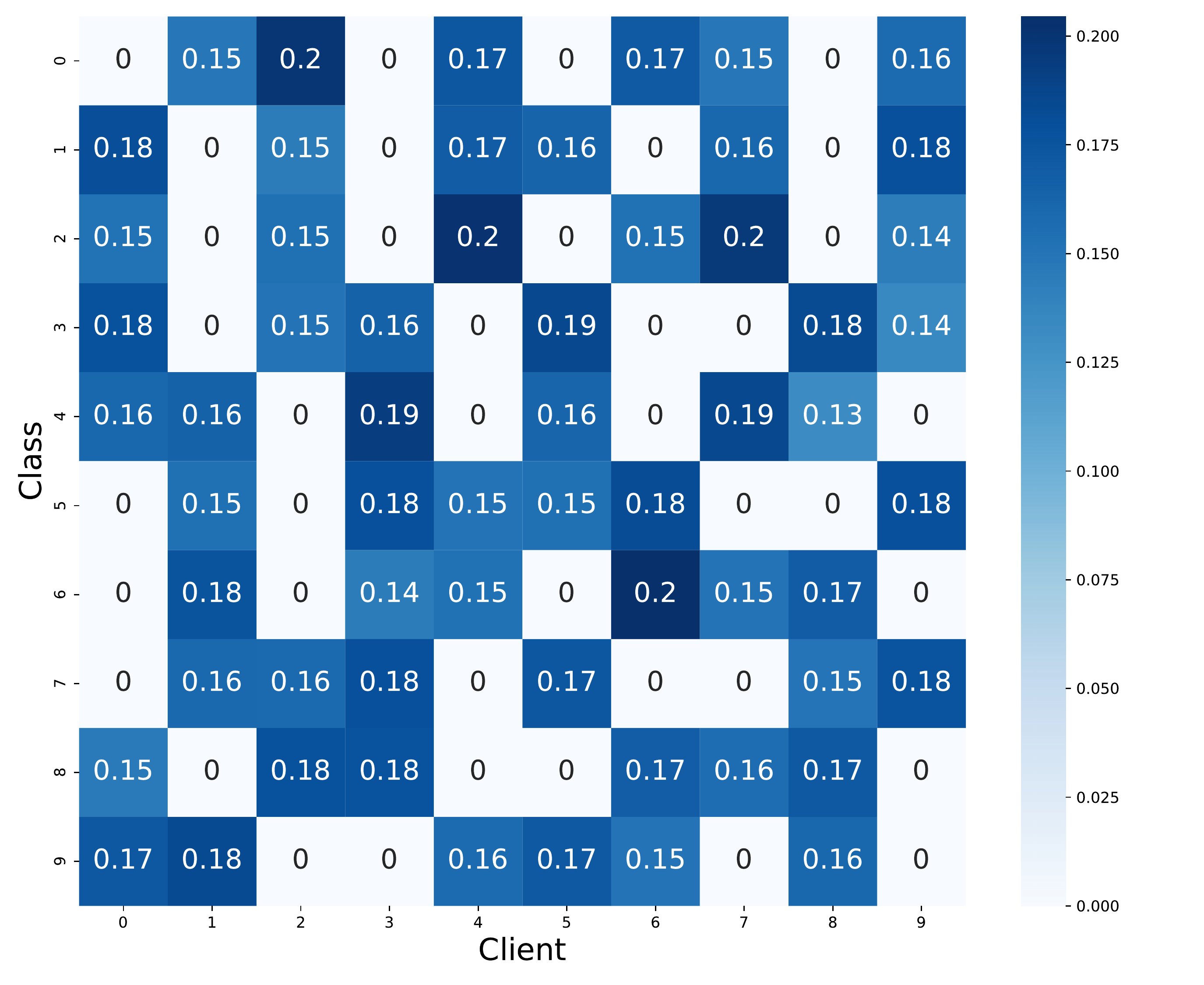}}
    \subcaptionbox{$DH=60\%$ }{\includegraphics[width=0.245\linewidth]{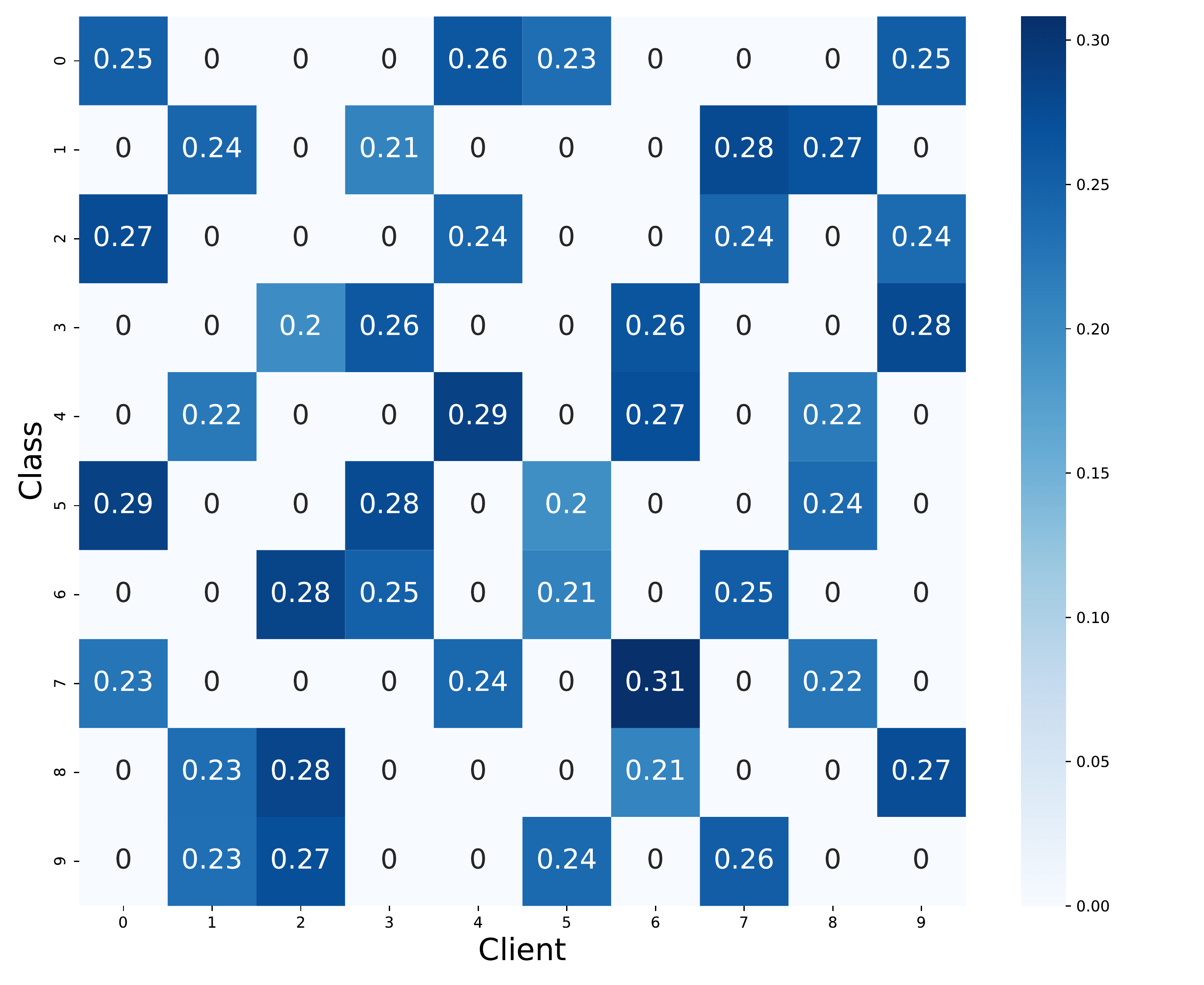}}
    \subcaptionbox{$DH=100\%$  }{\includegraphics[width=0.245\linewidth]{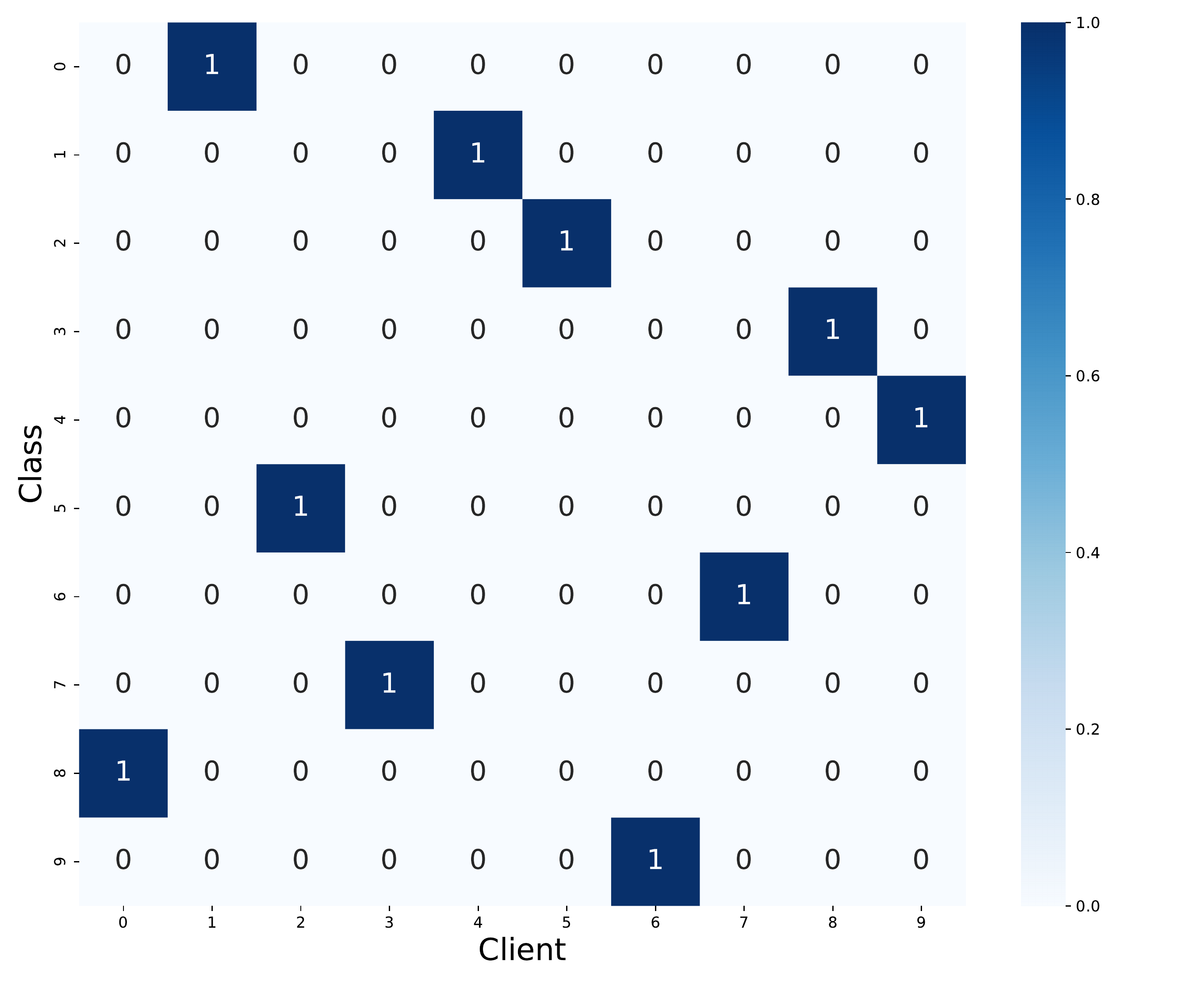}}

    \caption{Different distributional heterogeneity levels on CIFAR-10.} 

    \label{data_distribution}
\end{figure*}

\noindent{\bf Distributional Heterogeneity.}
We define distributional heterogeneity to characterize the class heterogeneity among the clients. Formally, we denote  distributional heterogeneity as $DH$ and define it as follows:
 \begin{equation}
    DH = 1- \frac{\sum_{j\in [1,N]}c_j}{N\times C},
\label{Data_heterogeneity}
\end{equation}
where $N$ is the total number of classes, $C$ is the total number of clients, and $c_{j}$ is defined as follows:
\begin{equation}
    c_{j}= \\
 \begin{cases}
     0, & \text{if only one client has data from class j,} \\
     k, & \text{if  $k > 1$ clients have data from class j.}
 \end{cases}\label{eq:indicator}
\end{equation}
%
%
%
  Our defined $DH$ has a value  between 0 and 100\%. In particular, $DH=0\%$ means that each client has data from the $C$ classes, e.g., the clients' local data are i.i.d., while $DH=100\%$ means that each class of data belongs to only one client, i.e., an extreme non-i.i.d. setting.  Figure~\ref{data_distribution} shows  examples of different levels of distributional heterogeneity  visualized by heatmaps for clients' local data in our experiments on CIFAR-10.  We list clients on the x-axis and  classes on the y-axis; and each cell is the fraction of the data from the corresponding class that are on the corresponding client. 

\subsection{Offset Aggregation Methods} 
\sys aggregates clients' offsets based on distributional heterogeneity. Intuitively, when the distributional heterogeneity is very large, the offset of one client may not be informative for the offset of another client, as their data distributions are substantially different. Therefore, we  aggregate clients' offsets only if the distributional heterogeneity is smaller than a threshold (we set the threshold to be 50\% in experiments). 


Suppose  the distributional heterogeneity of an FL system is smaller than the threshold. One naive way to aggregate the clients' offsets is to compute their average as a global offset, which is sent  back to all  clients. However, such naive aggregation method uses the same global offset for all clients, which achieves suboptimal accuracy as shown in our experiments. Therefore, we propose a neural network based aggregation method, which produces different aggregated offsets for the clients. Specifically, the server maintains a neural network, which takes a client-specific embedding vector ${e} \in \mathbb{R}^{1 \times N}$ and a client's offset as input and outputs an aggregated offset for the client,  where an entry $e_i$ of the embedding vector is the fraction of the training data in class $i$ that are on the client. 



The server learns the offset aggregation network by treating it as a regression problem during the FL training process. Specifically, in each round of \sys, the server collects a set of pairs ($t_i$, $t_i'$), where $t_i$ is the offset from client $i$ in the current round, $t_i'$ is the aggregated offset the server outputs for client $i$ in the previous round, and $i=1,2,\cdots,C$. The server learns the  offset aggregation network by minimizing the $\ell_2$ distance between $t_i$ and $t_i'$, i.e., $\min \sum_{i=1}^C ||t_i-t_i'||_2$, using Stochastic Gradient Descent (SGD). 

\section{Experiments} \label{Experiment}


\noindent{\bf Hyperparameters.}  Our model's architecture is the double-input-channel model as shown in Figure~\ref{fig:arch}. Our default $\alpha$ value is 0.3, number of epochs $E=1$, and the learning rates for the model and offset optimization are 5e-3 and 1e-3 respectively. Our neural network based offset aggregator's architecture is a single-input-channel generator with four convolutional layers.



	

        

        
        
        


\noindent{\bf Datasets and Model Architectures.} We use six different datasets in the experiment to show the generality of \sys. (i) The BioID~\cite{bioid} dataset contains 1521 gray level images with the frontal view of 23 people's face and eye positions. We keep 20 people's images in a descending order and central-crop the images into 256$\times$256.  (ii) The CelebA~\cite{celeb_A} dataset contains 202,599 face images of 10,177 unique, unnamed celebrities. Due to computation resource limit, we choose images of 50 identities in descending order, central-crop them to 178$\times$178, and then resize to 128$\times$128.  (iii) The CH-MNIST~\cite{ch_mnist} dataset contains eight classes of 
5,000 histology tiles images (64x64) from patients with colorectal cancer, (iv) The CIFAR-10~\cite{CIFAR-10} dataset contains 60,000 32$\times$32 color images in 10 different classes, we resize them to 64$\times$64, (v)The CIFAR-100~\cite{CIFAR-10} dataset contains 60,000 32$\times$32 color images in 100 different classes, and (vi) Caltech-UCSD Birds-200-2011~\cite{bird_200} (referred as Bird-200. The Bird-200 dataset contains 11,788 image from 200 bird species. Due to computation resource limit, we resize them to 128$\times$128. 

 Here are the model architectures for each dataset. 
 We use LeNet as the backbone for BioID, AlexNet for CelebA, CH-MNIST and CIFAR-100, ResNet18 and ResNet50 for CIFAR-100, and ResNet18 for Bird200.

\label{setup}
\noindent{\bf Local Data Distribution.}
 Our local data distribution ranges from 
 entirely i.i.d. to  extreme non-i.i.d., i.e., with distributional heterogeneity value ranging from 0\% to 100\%. 
 %
 %
  Our data splitting method follows  SOTA approach~\cite{hypernet}.  Specifically, 
  %
  we first assign a specific number of classes $u$ out of total classes $N$ for each client. Then, we sample $s_{i,c} \in (0.4, 0.6)$ for each client $i$ and a selected class $c$, and then assign the client with $\frac{s_{i,c}}{\sum_ns_{n,c}}$ of the samples for the class $c$.  We repeat the same process for each client.

\begin{table*}[!t]
\centering
\renewcommand{\arraystretch}{1.2} 
\setlength{\tabcolsep}{2pt}
 \scriptsize
\caption{\sys vs. SOTA under different data distribution. \NA\ means that the approach is not applicable under that setting, and DH means distributional heterogeneity (0\%: i.i.d. and 100\%: extreme non-i.i.d.).
 We did not evaluate the datasets of BioID and CelebA under other distributional settings due to the relative small number of images per class. 
 }
 
\label{vs.SOTA}
\begin{tabular}{c|c|c|c|c|c|c|c|c}
\toprule
Dataset & \# clients&  DH  & \sys (ours) & FedAvg& pFedMe &pFedHN & MOON &  FedAwS\cr
\midrule
\multirow{3}{*}{CH-MNIST} &\multirow{3}{*}{8}
& 0\% (i.i.d.) & \textbf{0.908} & 0.891& 0.778 & 0.702 & 0.887 & \NA \cr
&& 50\% & \textbf{0.907}       & 0.892 & 0.834 & 0.871 & 0.894 & \NA \cr
&&100\% & \textbf{0.946} & 0.908 & 0.908 & 0.641 & 0.910 & 0.942\cr
\midrule
\multirow{5}{*}{CIFAR-10} & \multirow{5}{*}{10} 
& 0\% (i.i.d.) & \textbf{0.829} & 0.809 & 0.520 & 0.652 & 0.789 &\NA \cr
&& 40\% & \textbf{0.819}          & 0.782 & 0.523 & 0.721 & 0.809 &\NA \cr
&& 60\% & \textbf{0.846}   & 0.751 & 0.673 & 0.785 & 0.798 &\NA \cr
& & 80\% & \textbf{0.891}  & 0.702 & 0.736 & 0.869 & 0.794 &\NA \cr
&& 100\%  & \textbf{0.860}  & 0.726 & 0.751 & 0.629 & 0.813 & 0.829 \cr
\midrule
\multirow{5}{*}{CIFAR-100} &  \multirow{5}{*}{10}
& 0\% (i.i.d.) & \textbf{0.533}     & 0.531 & 0.020 &0.354 & 0.532 &\NA \cr
&& 40\% & \textbf{0.586}           & 0.538 & 0.018 &0.492 & 0.564 &\NA \cr
&& 60\% & \textbf{0.646}           & 0.523 & 0.017 &0.604 & 0.628 &\NA \cr
&& 80\% & \textbf{0.734}           & 0.461 & 0.013 &0.669 & 0.709 &\NA \cr
&& 100\% & \textbf{0.834}           & 0.524 & 0.015 & 0.469& 0.820 &\NA \cr
\midrule
\multirow{5}{*}{Bird-200} &  \multirow{5}{*}{10}
& 0\% (i.i.d.) & \textbf{0.556}     & 0.518 & 0.018 &0.053 & 0.523 &\NA \cr
&& 40\% & \textbf{0.548}           & 0.521& 0.015 &0.064& 0.528 &\NA \cr
&& 60\% & \textbf{0.542}           & 0.528& 0.012 &0.086 &  0.532 &\NA \cr
&& 80\% & \textbf{0.565}           & 0.524& 0.010 &0.125 & 0.550 &\NA \cr
&& 100\% & \textbf{0.641} &0.549 & 0.014 & 0.309 & 0.621 &\NA \cr
\midrule

BioID & 20 & 100\%  & \textbf{0.988}  & 0.911 & 0.902 & 0.932 & 0.961 & 0.983 \cr
\midrule
CelebA & 50 & 100\% & \textbf{0.804} &0.639  & 0.527 & 0.545 & 0.497 &0.721 \cr
\bottomrule 
\end{tabular}
\end{table*}

\subsection{Results under Different Data Distributions}

We evaluate \sys's accuracy with different data distributions and compare with SOTA personalized FL works.

\noindent{\bf Extreme non-i.i.d.}
 The extreme non-i.i.d.\ setting, following prior work~\cite{FedAws}, is a setup where each client only has one class (called positive labels), thus being disjointed from each other. The distributional heterogeneity value is thus 100\%. We single out this setting, because
 the evaluation metrics are different from other settings given that each client only has positive images. That is, the same amount of negative images (i.e., randomly-selected images from other classes) are introduced in the testing dataset just like prior work~\cite{FedAws}.
 
 The rows with 100\% distributional heterogeneity values in Table~\ref{vs.SOTA} show the model's accuracy of \sys and the comparison with SOTA works. As shown in those results, \sys outperforms all prior works with five different datasets with an improvement ranging from 0.4\% to 7.7\%. FedAwS is clearly the SOTA, which always performs next to \sys, because it is designed for this extreme setting. Due to the negative test images, pFedHN performs poor since the server assigns each client model weights that are trained on only positive images according to its mechanism. FedAvg performs better than we expect because the features of negative examples are aggregated from other clients. We did not evaluate CIFAR-100 or Bird-200 under positive labels scenario (FedAws), since the number of classes per client does not satisfy positive labels setting when \# clients equals to 10 for them. Instead, each client is assigned 10 or 20 disjoint classes as the extreme non-i.i.d. case. 
 
 \noindent{\bf Other Distributional Settings.} Other settings include distributional heterogeneity values ranging from 0\% (i.i.d.) to 80\% .
  The evaluation also follows prior FL works~\cite{FedAvg,pFedMe}, i.e., each client evaluates testing data with the same classes as its training data.  
  Table~\ref{vs.SOTA} also shows the accuracy of \sys and four other SOTA works (FedAvg, pFedMe, MOON, and pFedHN). 
  \sys outperforms STOA works in every data distribution for all datasets. Note that we do not evaluate FedAwS in these settings because its design is only applicable to the extreme non-i.i.d.\ setting. 
  
  
  \subsection{Comparing with Data Transformation}
  
  We compare \sys with two state-of-the-art, popular data transformation (augmentation) methods, mixup~\cite{mixup} and AdvProp~\cite{augment_3}. The former, i.e., mixup, augments training data with virtual training data based on existing data samples and one hot encoding of the label. The latter, i.e., AdvProp, augments training data with its adversarial counterpart.  We add both data transformation methods for local training data at each client of FedAvg. 
  
  The comparison results are shown in Table~\ref{tab:aug}. \sys appears to outperform both mixup and AdvProp in different data distributions from i.i.d.\ to non-i.i.d. There are two major reasons. First, \sys shifts local training and testing data distribution to fit the global model, but existing data transformation only improves training data. Second, \sys aggregates the offset based on data distributions, but neither data transformation approaches did so.  Another thing worth noting is that mixup improves FedAvg under an i.i.d.\ setting, but AdvProp improves FedAvg under a non-i.i.d.\ setting. On one hand, that is likely because virtual examples under a non-i.i.d.\ setting may introduce further distributional discrepancies, while adversarial examples may help each local model better know the boundary. On the other hand, the distribution is the same under an i.i.d.\ setting and so does the virtual examples, but different adversarial examples may explore different boundaries at different clients.

   \begin{table}[!t]
  \setlength{\tabcolsep}{17.3pt}
  \captionsetup{font= footnotesize}
  \scriptsize
    \caption{Comparison with data transformation for CH-MNIST dataset.} 
    \label{tab:aug}
	\centering
	\begin{tabular}{c|ccccc}
		\toprule
        \multirow{2}{*}{Method} & \multicolumn{5}{c}{Distributional heterogeneity} \cr
        \cmidrule{2-6}
        & 0\% & 25\% &  50\% & 75\% & 100\% \cr
        \midrule
        FedAvg  & 0.891 & 0.893  & 0.892 & 0.847  & 0.908  \cr
         \sys  & \textbf{0.908} &\textbf{0.904}  & \textbf{0.907} & \textbf{0.905} & \textbf{0.946} \cr
         mixup  & 0.896 & 0.895 & 0.882  & 0.839 & 0.901  \cr
         AdvProp  & 0.879 & 0.880  & 0.877 & 0.859 & 0.919 \cr

        \bottomrule 
        
	\end{tabular}
\end{table}

\subsection{Ablation Studies}

        
        

\begin{table}[t!]
  \setlength{\tabcolsep}{9pt}
  \captionsetup{font= footnotesize}
  \scriptsize
    \caption{Ablation study on model structures. We adopt different distributional heterogeneity values according to the number of classes in the dataset, i.e., 0\%, 25\%, 50\%, 75\%, and 100\%  for CH-MNIST (8 classes) and 0\%, 40\%, 60\%, 80\%, and 100\% for CIFAR-10 (10 classes) and Bird-200 (200 classes). } 
    \label{tab:ab-double}
	\centering
	\begin{tabular}{c|c|ccccc}
		\toprule
        \multirow{2}{*}{Dataset} & \multirow{2}{*}{Structure} & \multicolumn{5}{c}{Distributional heterogeneity} \cr
        \cmidrule{3-7}
        & & 0\% & 40\%/25\% &60\%/50\% & 80\%/75\% & 100\% \cr
        \midrule
        \multirow{2}{*}{CH-MNIST} & single & 0.874 & 0.871   & 0.872  & 0.874 & 0.889 \cr
        & double & \textbf{0.908}  & \textbf{0.904} & \textbf{0.907}  & \textbf{0.905}  & \textbf{0.946}\cr
        
        \midrule
        \multirow{2}{*}{CIFAR-10} & single & 0.775 & 0.802 & 0.785 & 0.796 & 0.811  \cr
        & double  & \textbf{0.829} & \textbf{0.819}  & \textbf{0.846} & \textbf{0.891} &\textbf{ 0.860}\cr
        \midrule
        \multirow{2}{*}{Bird-200} & single & 0.569 & 0.512& 0.497 & 0.501 & 0.505\cr
        & double  & \textbf{0.556} & \textbf{0.548}  & \textbf{0.542}  & \textbf{0.565} & \textbf{0.641} \cr
        
        \bottomrule 
        
	\end{tabular}
\end{table}

\noindent{\bf Single vs. Double-Input Channel.}
 We compare the performance of single vs. double-input-channel models to demonstrate the necessity in using the double-input-channel model. Table~\ref{tab:ab-double} shows the model's accuracy on three datasets with different distributional heterogeneity values. As shown, the double-input-channel model always outperforms the single-input-channel with around 3\%--9\% accuracy improvement on three different datasets.

\begin{table}[!t]
  \setlength{\tabcolsep}{7pt}
  \captionsetup{font= footnotesize}
  \scriptsize
    \caption{Ablation study on aggregation methods. We adopt different distributional heterogeneity values according to the number of classes in the dataset, i.e., 0\%, 25\%, 50\%, 75\%, and 100\%  for CH-MNIST (8 classes) and 0\%, 40\%, 60\%, 80\%, and 100\% for CIFAR-10 (10 classes) and Bird-200 (200 classes).} 
    \label{tab:ab-agg}
	\centering
	\begin{tabular}{c|c|ccccc}
		\toprule
        \multirow{2}{*}{Dataset} & \multirow{2}{*}{Aggregation} & \multicolumn{5}{c}{Distributional heterogeneity} \cr
        \cmidrule{3-7}
        & & 0\% & 40\%/25\% &60\%/50\% & 80\%/75\% & 100\% \cr
        \midrule
        \multirow{4}{*}{CH-MNIST} & no agg  & 0.868 & 0.887  & {0.907}  & {0.905}  & {0.946}\cr
        & avg agg  & 0.903  & 0.902  &{0.907}  & 0.865 &  0.899 \cr
        & nn agg   & {0.908} & {0.904}  & 0.905  & 0.887 &  0.921 \cr
        \cmidrule{2-7}
        & nn+no (default)   & \textbf{0.908} & \textbf{0.904}  & \textbf{0.907}  & \textbf{0.905} &  \textbf{0.946} \cr
        \midrule
        \multirow{4}{*}{CIFAR-10} & no agg & 0.767 & 0.789   & {0.846} & {0.891} & {0.860}\cr
        & avg agg  & 0.811 & 0.814   & 0.813  & 0.702 &  0.798 \cr
        & nn agg  & {0.829} & {0.819}  & 0.799   & 0.743 &  0.839 \cr
        \cmidrule{2-7}
        & nn+no (default)  & \textbf{0.829} & \textbf{0.819} & \textbf{0.846} & \textbf{0.891} & \textbf{0.860}\cr
        \midrule
        \multirow{4}{*}{Bird-200} & no agg & 0.501 & 0.522   & 0.542 & {0.565} & {0.641}\cr
        & avg agg  & 0.526 & 0.529   & 0.525 & 0.515  &  0.489\cr
        & nn agg   & {0.556} & {0.548}   &  {0.551} & 0.532 & 0.513  \cr
        \cmidrule{2-7}
        & nn+no (default) & \textbf{0.556} & \textbf{0.548} & \textbf{0.551} & \textbf{0.565} & \textbf{0.641}\cr
        \bottomrule 
	\end{tabular}
\end{table}

\noindent{\bf Different Offset Aggregations.}
 We compare different offset aggregation methods, i.e., no aggregation, average aggregation and neural network (NN) based aggregation, on three datasets with various distributional heterogeneity values. 
 Table~\ref{tab:ab-agg} shows the comparison results. No aggregation performs best when the distributional heterogeneity is greater than 50\%, and NN aggregation performs the best when the distributional heterogeneity is smaller than 50\%. Average aggregation always performs worse than the other two.  This motivates the design of \sys in adopting no aggregation for greater than 50\% distributional heterogeneity and NN aggregation for less than 50\% distributional heterogeneity, which is the ``nn+no (default)'' row in Table~\ref{tab:ab-agg}.



\begin{figure}[!t]
\centering
\begin{minipage}[t]{0.32\textwidth}
\centering
\includegraphics[width=3.7cm]{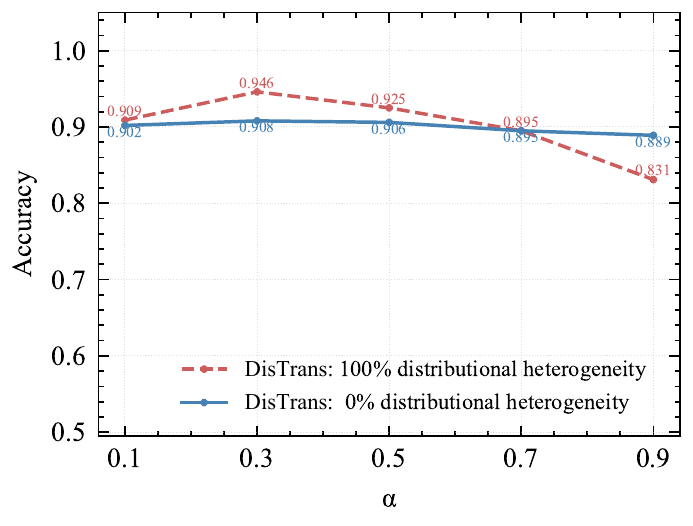}
\caption{Accuracy vs. $\alpha$ for CH-MNIST.}\label{fig:ablation}

\end{minipage}
\hfill
\begin{minipage}[t]{0.32\textwidth}
\centering
\includegraphics[width=3.7cm]{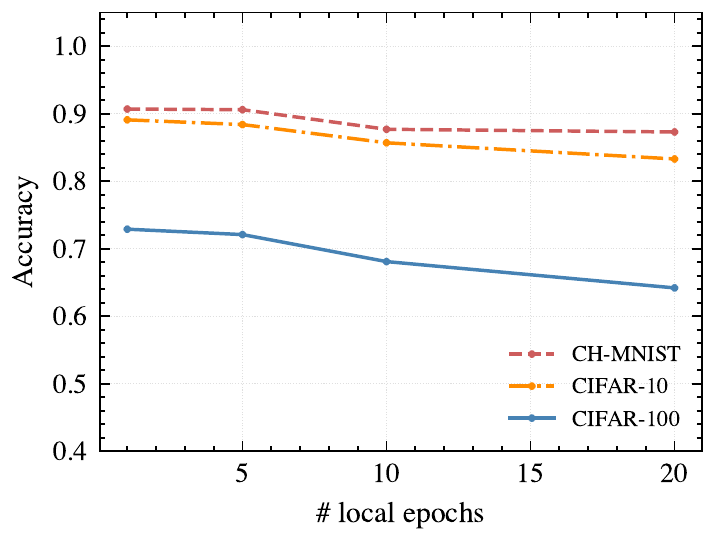}
\caption{{Accuracy vs. \# of local epochs.}}\label{fig:ablation-epoch} 
   
\end{minipage}
\hfill
\begin{minipage}[t]{0.32\textwidth}
\centering
\includegraphics[width=3.7cm]{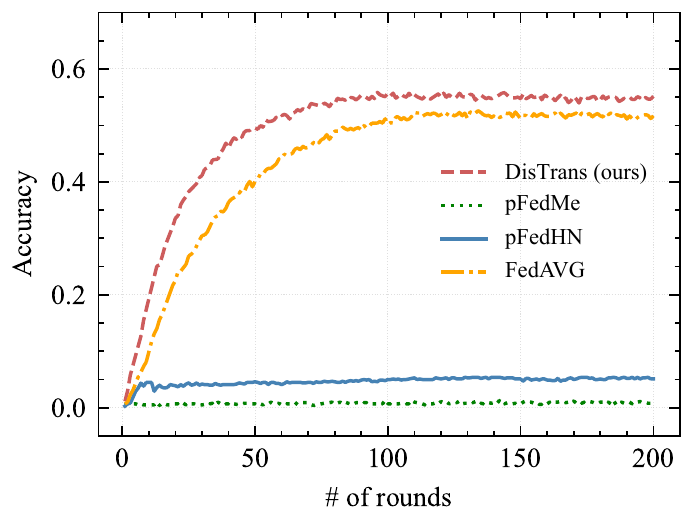}
\caption{Accuracy vs. \# of rounds for Birds-200.
   }\label{fig:rounds} 
   
\end{minipage}
\end{figure}



\noindent{\bf Different $\alpha$ Values.}
 We evaluate top-1 accuracy of \sys with different $\alpha$ values, and 0\% and 100\% distributional heterogeneity to justify why we choose 0.3 as $\alpha$. Figure~\ref{fig:ablation} shows the results.
  The accuracy with 100\% distributional heterogeneity is more sensitive to $\alpha$ than that with 0\%. In both data distributions, the accuracy is the highest when $\alpha$ equals 0.3.  The reason is as follows.  When $\alpha$ is small, the \trigger is too weak to shift the distribution. When the $\alpha$ is large, the \trigger is too strong in overriding the original data distribution. 

\noindent{\bf {Different Local Epochs.}}
{We study different local training epochs for each round. Figure~\ref{fig:ablation-epoch} shows the accuracy for CH-MNIST, CIFAR-100, and Bird-200 with epochs from 1, 5, 10, to 20. The accuracy is the highest with the local epoch as 1, and decreases when the epoch increases. The reason is too much local training makes offsets become overfitted to local data.} 

\noindent{\bf Convergence Rate.}
 We study the convergence rate of three SOTA works and \sys in terms of communication rounds between the server and local clients. Figure~\ref{fig:rounds} shows the number of communication rounds as the x-axis and the model's accuracy as the y-axis for the Birds-200 dataset under the i.i.d.\ setting (i.e., 0\% distributional heterogeneity).   There are two things worth noting. First, as shown, the convergence rate of \sys is similar to that of FedAvg, which needs approximately 100 rounds. Second, the accuracy of \sys is constantly better than FedAvg for each communication between client and server. 

\subsection{Scalability} 

\begin{table}[t!]
  \setlength{\tabcolsep}{5.5pt}
  \scriptsize
    \caption{Best accuracy and the number of rounds to achieve it vs. different number of clients using the CH-MNIST dataset when reaching listed accuracy, e.g., \sys needs 5 rounds to achieve a 0.800 accuracy with 8 clients. (\NA: the approach cannot reach the accuracy under that setting.)} 
    \label{tab:scale-2}
	\centering
	\begin{tabular}{c|c|c|ccccc}
		\toprule
		& \multirow{2}{*}{\# clients} & \multirow{2}{*}{Best accuracy} & \multicolumn{5}{c}{\# of rounds to achieve} \cr
         &&& $>$0.700 & $>$0.800 & $>$0.850  & $>$0.870 & $>$0.890  \cr
        \midrule
          \multirow{4}{*}{\sys(ours)} & 8  
                                      &\textbf{0.907}& \textbf{2} & \textbf{5} & \textbf{10} & 36 & \textbf{63} \cr
                                      & 16 &\textbf{0.898} & 8 & 15 & 25 & 51 & \textbf{123} \cr
                                      & 24 &\textbf{0.897} & 12 & 26 & \textbf{37} & \textbf{68} & \textbf{154}\cr
                                      & 40 &\textbf{0.895} & 14 & 29 & \textbf{40} & \textbf{87} & \textbf{192}\cr
        \midrule
                                      
        \multirow{4}{*}{FedAVG} & 8 &0.892 & 3& \textbf{5}&  15&  \textbf{24}& 78  \cr
                                      & 16 &0.883 & 7 &  \textbf{12}&  \textbf{23} &  \textbf{48}& \NA \cr
                                      & 24 & 0.880 & 10 & \textbf{21} & 39 & 74 & \NA \cr
                                      & 40 & 0.878 & 19&  34&   44&  100 & \NA \cr
        \midrule                                  
        \multirow{4}{*}{pFedMe} & 8 &0.834 &  690 & 779 & \NA & \NA & \NA \cr
                                      & 16 &0.805 & 844 & 1,225 & \NA & \NA & \NA \cr
                                      & 24 &0.725 & 1,859 & \NA & \NA & \NA & \NA \cr
                                      & 40 &0.719 & 3,071 & \NA &  \NA & \NA  & \NA \cr
        \midrule                                  
        \multirow{4}{*}{pFedHN} & 8 &0.871 & 3&  8&  23& 117 & \NA \cr
                                      & 16 &0.817 & \textbf{6} & 29 & \NA & \NA & \NA \cr
                                      & 24 &0.816 & \textbf{4}&  28& \NA & \NA & \NA  \cr
                                      & 40 &0.832 & \textbf{5}& \textbf{27} &  \NA&  \NA & \NA \cr

        \bottomrule 
	\end{tabular}
	\end{table} 

We study the scalability of \sys using two datasets CH-MNIST and CIFAR-100 as the number of FL clients increases. First, we test the number of clients from 8, 16, 24, to 40 using 50\% distributional heterogeneity for CH-MNIST. The third column in Table~\ref{tab:scale-2} shows the accuracy of four different works including \sys as the number of clients increases. 
 The fourth to eighth columns in Table~\ref{tab:scale-2} show the total number of rounds to reach certain accuracy.  Generally, the convergence needs more rounds with more clients, which aligns with the previous work~\cite{FedAvg}. 
 Second, we show the testing accuracy in Table~\ref{tab: scala-cifar} for CIFAR-100 (with ResNet18) of 50, 100, and 500 clients. Each client has 10 classes and the sample rate is 0.2~\cite{moon}. 
 \sys outperforms SOTA by 8.8\%--30.7\%. Note that the accuracy of \sys drops by 8.4\% while SOTA drops by 10.8\% to 27.6\% for 500 clients.




\begin{table}[h]
\centering
\renewcommand{\arraystretch}{1} 
\setlength{\tabcolsep}{2.8pt}
\scriptsize
\caption{{Best accuracy vs. number of clients on CIFAR-100.}}
\label{tab: scala-cifar}
\begin{tabular}{c|ccc|ccc|ccc|ccc}
\toprule
 &
 \multicolumn{3}{c|}{\sys}& \multicolumn{3}{c|}{pFedHN} & \multicolumn{3}{c|}{pFedHN-pc} & \multicolumn{3}{c}{MOON} \cr
\midrule
\#Client&
50 & 100 & 500& 50 & 100 & 500  & 50 & 100 & 500 & 50 & 100 & 500\cr
\midrule
Accuracy &\textbf{0.729}  & \textbf{0.681} &\textbf{ 0.645 } &0.614 &0.538 & 0.338 & 0.623 & 0.541 & 0.372 & 0.615 & 0.593 & 0.507  \cr
\bottomrule
\end{tabular}
\end{table}



\begin{table}[!t]
  \setlength{\tabcolsep}{6pt}
  \scriptsize
    \caption{Communication overhead of weights and \trigger for 64x64 RGB images.} 
    \label{tab:communi}
	\centering
	\begin{tabular}{c|c|cc|c}
		\toprule
          & Baseline (in bytes) & \multicolumn{2}{c|}{\sys (in bytes)}  & \multirow{2}{*}{$\Delta$ overhead}\cr
          \cmidrule{2-4}
          & single-input-channel weight & double-input-channel weight & \trigger & \cr
        \midrule
        LeNet & 4,346,447 & 4,350,415 & 49,280 & 1.225\%\cr
        AlexNet & 244,449,263 & 244,489,263  & 49,280 & 0.036\%\cr
        ResNet18 & 44,805,709 & 44,826,189  & 49,280 & 0.155\%\cr
        {ResNet50} & 94,326,992 & 94,408,912  & 49,280 & 0.139\%\cr
        \bottomrule 
	\end{tabular}
\end{table}

\subsection{Communication Overhead}


We study the communication overhead by calculating the $\Delta$ bytes brought by our double-input-channel weights and \trigger in each communication round. The comparison baseline used is FedAvg with conventional single-input-channel model. 
Table~\ref{tab:communi} shows  results for different backbone neural network architectures: The overhead is between 0.036\% to 1.225\% with an average value of 0.389\% for different network architectures because the double-input-channel model only introduces one additional layer and the offsets are small. 
%
%


\subsection{Prediction Visualization}

\begin{figure}[!t]
    \subcaptionbox{FedAvg\label{fig:umapfedavg}}{\includegraphics[trim=0 50 0 0,width=0.245\linewidth]{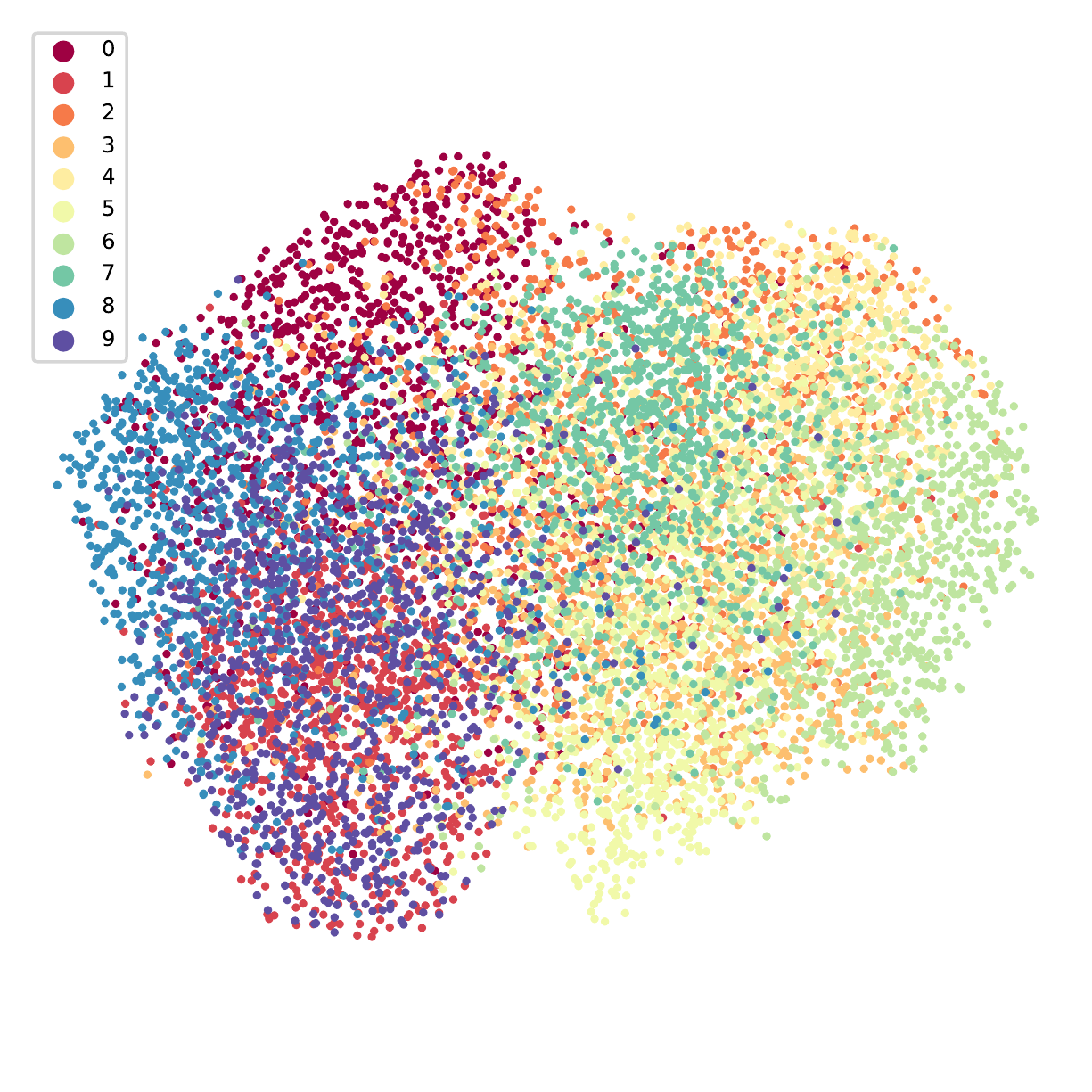}} 
    \subcaptionbox{FedAwS\label{fig:umapfedaws}}{\includegraphics[trim=0 50 0 0,width=0.245\linewidth]{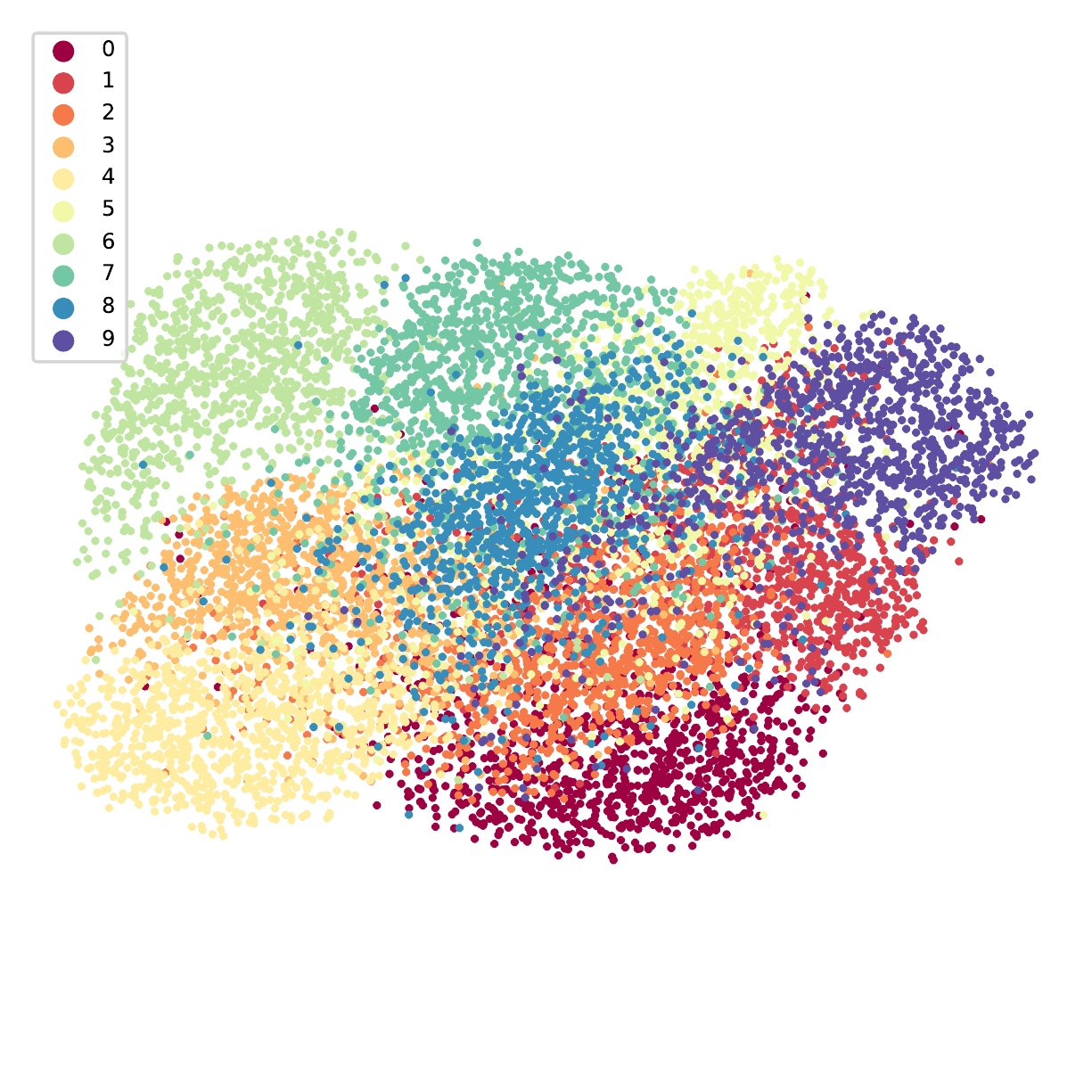}} 
    \subcaptionbox{pFedMe\label{fig:umappfedme}}{\includegraphics[trim=0 50 0 0,width=0.245\linewidth]{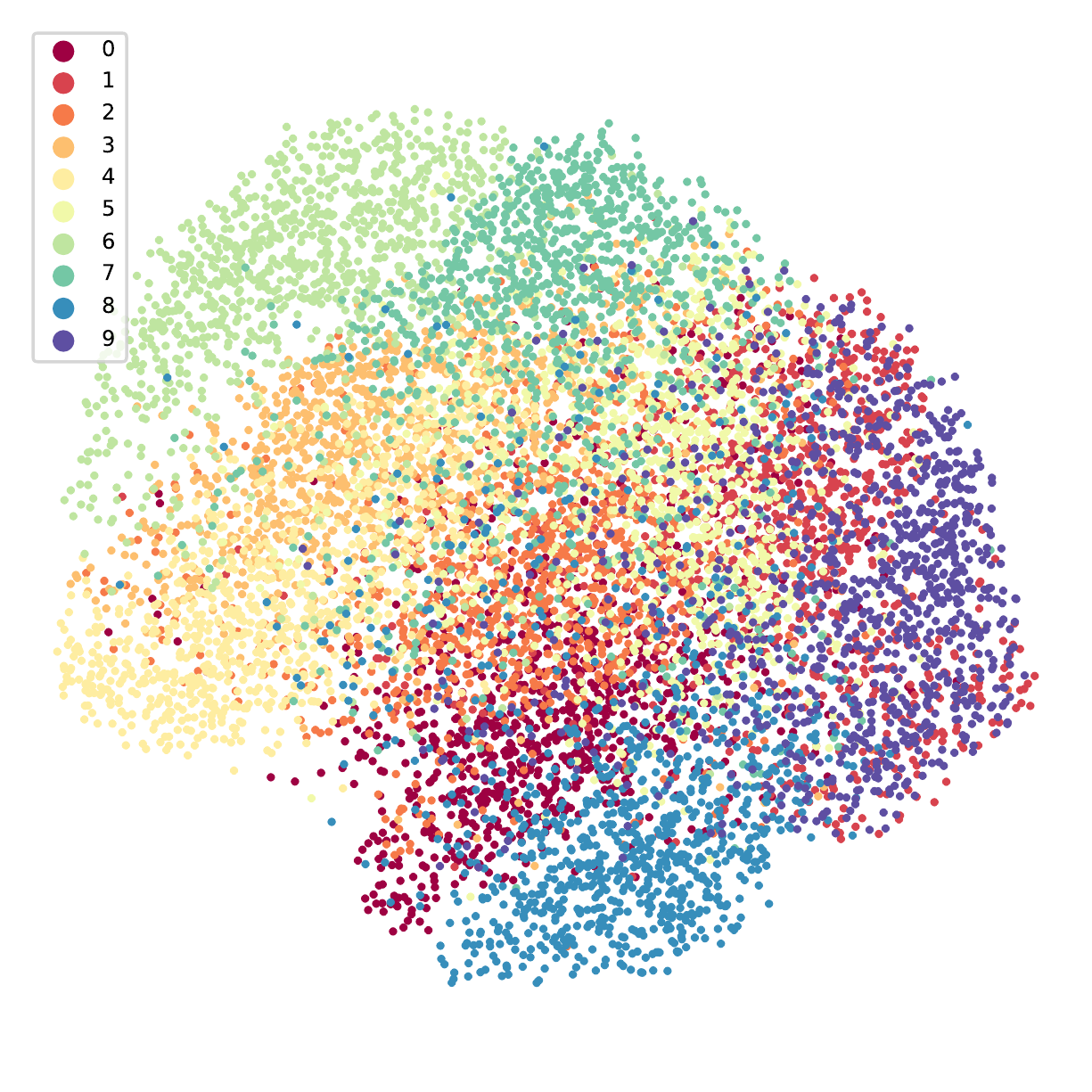}} 
    \subcaptionbox{\sys (ours) \label{fig:umap2}}{\includegraphics[trim=0 50 0 0,width=0.245\linewidth]{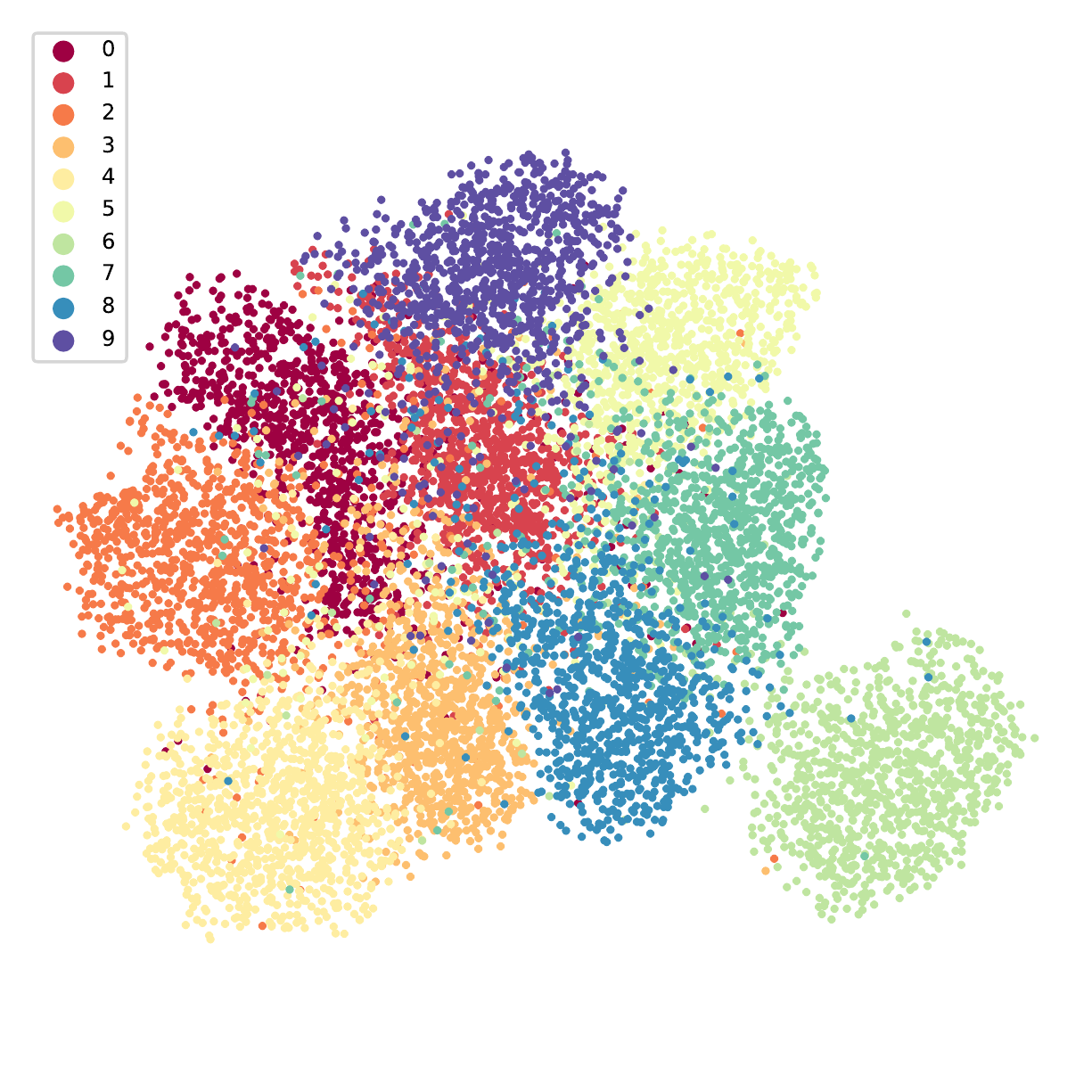}}
    \caption{UMAP visualization of embedded feature representations in the global model for test images in CIFAR-10. 
      \sys learns  better feature representations than FedAvg, FedAwS, and pFedMe. }
    \label{fig:umap} 
\end{figure}


We perform an experiment on CIFAR-10 using ten FL clients where each client has data for only one class, and visualize hidden feature representations using Uniform Manifold Approximation and Projection (UMAP) in   Figure~\ref{fig:umap}. 
The model trained using FedAvg  learns poor features, which are mixed and indistinguishable. 
The feature representations of FedAwS and pFedMe also highly overlap. By contrast, the feature representations of \sys are well separated in Figure~\ref{fig:umap2} as a result of  shifting the local data distributions via personalized offsets. 

\section{Conclusion}

FL often needs to contend with client-side local training data with different distributions  with high heterogeneity. This paper advances a novel approach, \sys, based on distributional transformation, that jointly optimizes local model and data with a personalized offset and then aggregates both at a central server. We perform an empirical evaluation of \sys using five different datasets, which shows that \sys outperforms SOTA  FL  and data augmentation methods, under different degrees of data distributional heterogeneity ranging from extreme non-i.i.d.\ to i.i.d.

\section*{Acknowledgements}

 This work was supported
in part by 
Johns Hopkins
University Institute for Assured Autonomy (IAA) with grants 80052272 and 80052273, and National Science Foundation (NSF) under grants
CNS-21-31859, CNS-21-12562, and CNS-18-54001.
 The views and conclusions contained
herein are those of the authors and should not be interpreted as
necessarily representing the official policies or endorsements,
either expressed or implied, of NSF or JHU-IAA.

%
%

\bibliographystyle{splncs04}
\bibliography{egbib}
\end{document}